\documentclass[10pt,twocolumn,letterpaper]{article}

\usepackage{cvpr}
\usepackage{times}
\usepackage{epsfig}
\usepackage{graphicx}
\usepackage{amsmath}
\usepackage{amssymb}

\usepackage[breaklinks=true,bookmarks=false]{hyperref}

\cvprfinalcopy 

\def\cvprPaperID{****} 

\begin{document}

\title{Deep Action- and Context-Aware Sequence Learning for Activity Recognition and Anticipation}

\author{Mohammad Sadegh Aliakbarian$^{1,2}$, Fatemehsadat Saleh$^{1,2}$, Basura Fernando$^{1}$, Mathieu Salzmann$^{3}$, \\Lars Petersson$^{1,2}$, Lars Andersson$^{2}$\\ \\
$^{1}$Australian National University, $^{2}$Smart Vision Systems, CSIRO, $^{3}$CVLab, EPFL\\
{\tt\small firstname.lastname@data61.csiro.au}, {\tt\small basura.fernando@anu.edu.au}, {\tt\small mathieu.salzmann@epfl.ch}
}

\maketitle

\begin{abstract}
Action recognition and anticipation are key to the success of many computer vision applications. Existing methods can roughly be grouped into those that extract global, context-aware representations of the entire image or sequence, and those that aim at focusing on the regions where the action occurs. While the former may suffer from the fact that context is not always reliable, the latter completely ignore this source of information, which can nonetheless be helpful in many situations. In this paper, we aim at making the best of both worlds by developing an approach that leverages both context-aware and action-aware features. At the core of our method lies a novel multi-stage recurrent architecture that allows us to effectively combine these two sources of information throughout a video. This architecture first exploits the global, context-aware features, and merges the resulting representation with the localized, action-aware ones. Our experiments on standard datasets evidence the benefits of our approach over methods that use each information type separately. We outperform the state-of-the-art methods that, as us, rely only on RGB frames as input for both action recognition and anticipation.


\end{abstract}

\begin{figure*}
\centering
\includegraphics[width=0.95\textwidth]{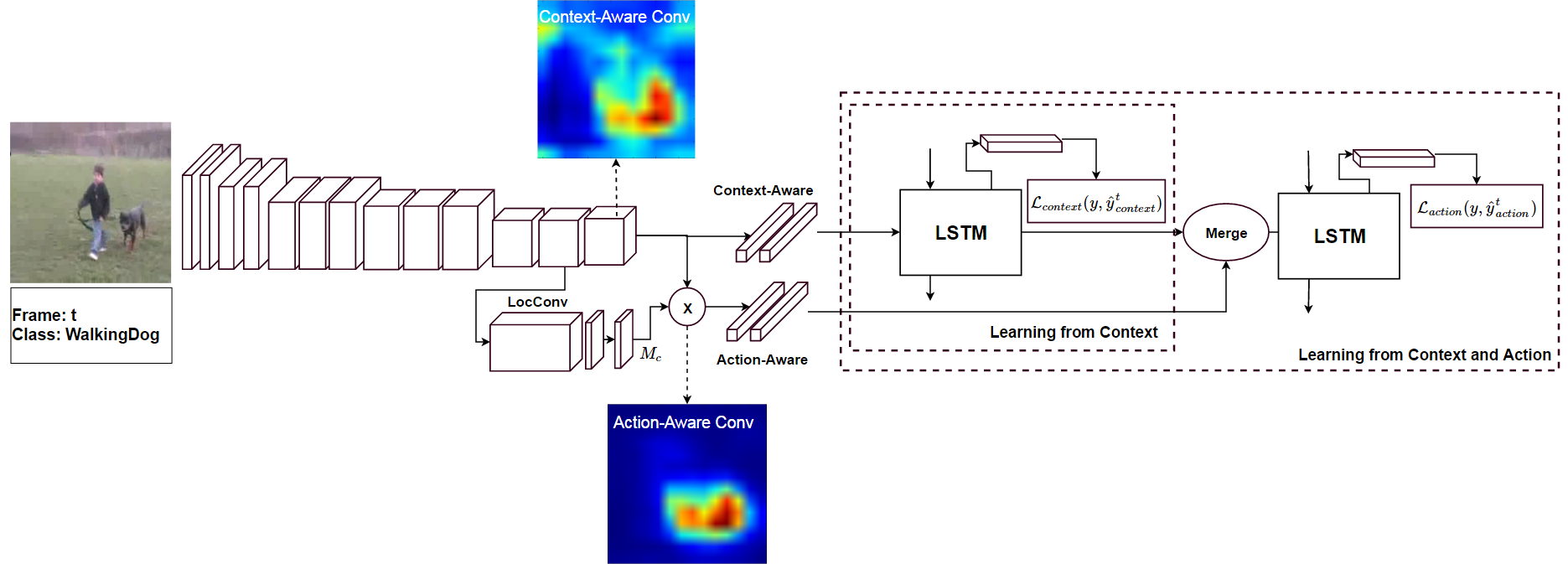}
\caption{{\bf Overview of our approach.} We propose to extract context-aware features, encoding global information about the scene, and combine them with action-aware ones, which focus on the action itself. To this end, we introduce a multi-stage LSTM architecture that leverages the two types of features to predict the action or forecast it. Note that, for the sake of visualization, the color maps were obtained from 3D tensors (\(512\times W\times H\)) via an average pooling operation over the 512 channels.
}
\label{FIG:LSTM}
\end{figure*}

\section{Introduction}
\label{SEC:INTRODUCTION}
Activity recognition and anticipation are crucial for the success of many real-life applications, such as autonomous navigation, sports analysis and personal robotics. It has therefore become increasingly popular in the computer vision literature. Nowadays, the most popular trend to tackle these tasks consists of extracting global representations for the entire image~\cite{DeepCAMP,TSN,ActionTransformation,LRCN}, or video sequence~\cite{3DCNN,LargeScaleCNN}. As such, these methods do not truly focus on the actions of interest, but rather compute a \emph{contex-aware} representation. Unfortunately, context does not always bring reliable information about the action. For example, one can play guitar in a bedroom, a concert hall or a yard. Therefore, the resulting representations encompass much irrelevant noise.

By contrast, several methods have attempted to localize the feature extraction process to regions of interest. This, to some degree, is the case of methods exploiting dense trajectories~\cite{TrajectoryPooled,IDT,DiscriminativeRankPooling} and optical flow~\cite{CNN2Stream,TSN,VLAD3}. By relying on motion, however, these methods can easily be distracted by irrelevant phenomena such as moving background or camera. Inspired by objectness, the notion of actionness~\cite{actionness,ActionnessRanking,ActionTubelets,FastActionProposal,OnlineSEEDS,SpatioTemporalProposal} has recently been proposed as a means to overcome this weakness by attempting to localize the regions where a generic action occurs. The resulting methods can then be thought of as extracting \emph{action-aware} representations. In other words, these methods go to the other extreme and completely discard the notion of context. In many situations, however, context provides helpful information about the action class. For example, one typically plays soccer on a grass field.

In this paper, we propose to make the best of both world: We introduce an approach that leverages both context-aware and action-aware features for action recognition and anticipation. In particular, we make use of the output of the last layer of an image-based Convolutional Neural Network (CNN) as context-aware features. For the action-aware ones, inspired by the approach of~\cite{zhou2015learning} for object recognition and localization, we propose to exploit the class-specific activations of another CNN, which typically correspond to regions where the action occurs. The main challenge then consists of effectively leveraging the two types of features for recognition and anticipation. To this end, we introduce the novel multi-stage recurrent architecture depicted by Fig.~\ref{FIG:LSTM}. In a first stage, this model focuses on the global, context-aware features, and combines the resulting representation with the localized, action-aware ones to obtain the final prediction. In short, it first extracts the contextual information, and then merges it with the localized one.

To the best of our knowledge, our work constitutes the first attempt at \emph{explicitly} bringing together these two types of information for action recognition.
By leveraging RBG frames and optical flow, the two-stream approach of~\cite{TwoStreamNIPS} exploits context and motion. As mentioned above, however, motion does not always correlate with the action of interest. While 3D CNNs~\cite{3DCNN} can potentially implicitly capture information about both context and action, they are difficult to train and computationally expensive. By contrast, our novel multi-stage LSTM model explicitly combines these two information sources, and provides us with an effective and efficient action recognition and anticipation framework.

As a result, our approach outperforms the state-of-the-art methods that, as us, rely only on RBG frames as input on all the standard benchmark datasets that we experimented with, including UCF-101~\cite{UCF101} and JHMDB21~\cite{JHMDB}. Furthermore, our experiments clearly evidence the benefits of our multi-stage architecture over networks that exploit either context-aware or action-aware features separately, or combine them via other fusion strategies.

\section{Related Work}
\label{SEC:RELATED}

Over the years, great progress has been made in activity recognition~\cite{SpaceTimeInterest,IDT,LRCN,DiscriminativeRankPooling,MultiStream,TSN,MultiStreamBiLSTM}. 
Unsurprisingly, while earlier approaches relied on handcrafted features~\cite{IDT,SpaceTimeInterest}, recent ones have turned towards deep learning. Below, we focus on these approaches, which are most related to our work.


In this deep learning context, many methods rely on CNNs~\cite{3DCNN,LargeScaleCNN,CNN2Stream,CovAction,MutiRegion2Stream} to extract a global representation of images. These CNN-based methods, however, typically have small temporal support, and thus fail to capture long-range dynamics. For instance, the two-stream networks~\cite{TSN,CNN2Stream,TwoStreamNIPS} act on single images in conjunction with optical flow information to model the temporal information. While 3D convolutional filters have been proposed~\cite{3DCNN}, they are typically limited to acting on small sets of stacked video frames, 10 to 20 in practice.

By contrast, recurrent architectures, such as the popular Long-Short Term Memory networks~\cite{LSTM}, can, in principle, learn complex, long-range dynamics, and have therefore recently been investigated for action recognition~\cite{LRCN,RegularizeSkeleton,LSTMAction,MultiStreamBiLSTM,SpatioTemporalLSTM,LSTMAction}. For instance, in~\cite{LRCN}, an LSTM was employed to model the dynamics of CNN activations; in~\cite{MultiStreamBiLSTM}, a bi-directional LSTM was combined with a multi-stream CNN to encode the long-term dynamics within and between activities in videos. Other works, such as~\cite{RegularizeSkeleton}, have proposed to exploit additional annotations, in the form of 3D skeletons, into an LSTM-based model. Such annotations, however, are not always available in practice, thus limiting the applicability of these methods.

Beyond recurrent models, rank pooling has also proven effective to model activities in videos~\cite{RankPooling,DynamicNetwork,DiscriminativeRankPooling}. In this context,~\cite{RankPooling} computes a representation encoding the dynamics of the video, and~\cite{DynamicNetwork} introduces the concept of Dynamic Images to summarize the gist of a sequence.

In any event, whether based on CNNs, LSTMs or rank pooling, all of the above-mentioned methods compute one holistic representation over one image, or the sequence. While this has the advantage of retaining information about the context of the action, these methods may also easily be affected by the fact that context is not always reliable. Many actions can be performed in very different environments. In these cases, focusing on the action itself would therefore seem beneficial.

This, in essence, is the goal of methods based on the notion of actionness~\cite{actionness,ActionnessRanking,ActionTubelets,FastActionProposal,OnlineSEEDS,SpatioTemporalProposal}. Inspired by the concept of objectness~\cite{objectness,selectiveSearch}, commonly used in object detection, actionness aims at localizing the regions in a video where an action of interest occurs. In~\cite{actionness}, this was achieved by exploiting appearance (RGB) and motion (optical flow) in a two-stream architecture. In~\cite{actionness}, the resulting actionness map was then employed to generate action/bounding box proposals via an action detection framework based on~\cite{RCNN}, and classifying these proposals. The ActionTube approach of~\cite{actionTube} follows a similar framework, but relies on~\cite{FRCNN} instead of~\cite{RCNN}. More importantly, by focusing on the actions themselves, these methods throw away all the information about context. However, in many scenarios, such as to recognize different sports, context provides helpful information about the observed actions. Note that for extracting actionness in~\cite{actionness,actionTube}, bounding box annotation are used as an extra supervision during the training process, while our approach requires no additional annotations.

In short, while one class of methods model images in a global manner, and may thus be sensitive to context diversity, the other ones focus solely on the action, and thus cannot benefit from context. Here, we introduce a novel multi-stage recurrent architecture that explicitly and effectively combines these two complementary information sources.

\section{Preliminary}
\label{SEC:PRELIMINARY}
Here, we briefly talk about the main building block of our architecture, the LSTM. LSTM is a neural network that implements a memory cell which can maintain its state over time. Hence, the benefit of LSTM units is that they allow the recurrent network to remember long-term context dependencies and relations~\cite{LSTM}.

LSTM consists of three gates: (1) input gate $i$, (2) output gate $o$, and (3) forget gate $f$ – and a memory cell $c$. At each time-step $t$, LSTM first computes the activation of its gates and then updates its memory cell from $c_{t−1}$ to $c_t$. It then computes the activation of the output gate $o_t$, and finally outputs a hidden representation $h_t$. There are two inputs to the LSTM that each time-step: (1) the observations $x_t$ and (2) the hidden representation from the previous time step $h_{t−1}$. To update, LSTM applies the following equations:
\begin{equation}
i_t = \sigma(W_ix_t + U_ih_{t−1} + V_ic_{t−1} + b_i)
\end{equation}
\begin{equation}
f_t = \sigma(W_fx_t + U_fh_{t−1} + V_fc_{t−1} + b_f)
\end{equation}
\begin{equation}
c_t = f_t \bullet c_{t−1} + i_t  tanh(W_cx_t + U_ch_{t−1} + b_c)
\end{equation}
\begin{equation}
o_t = \sigma(W_ox_t + U_oh_{t−1} + V_oc_t + b_o)
\end{equation}
\begin{equation}
h_t = \sigma_t  tanh(c_t)
\end{equation}

To update the memory of the LSTM, input gate and forget gate will be involved. In more detail, what an input gate do is computing new values based on new observations that are going to be written in the memory cell and the forget gate participate in a part of memory cell to forget. Both output gate and memory are responsible for computing the representation of hidden units. What makes the gradient of LSTM gets propagated over a longer time before vanishing is that LSTM activations contain summation over time and also derivatives are distributed over the summations.

\begin{figure}
\centering
\includegraphics[width=0.45\textwidth]{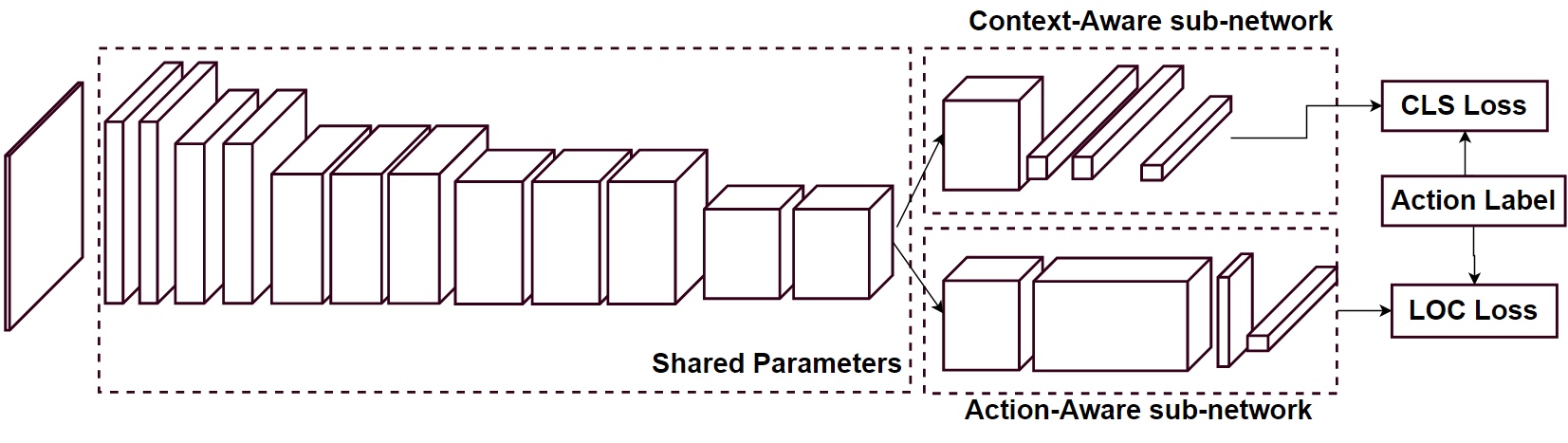}
\caption{{\bf Our feature extraction network.} Our CNN model for feature extraction is based on the VGG-16 structure with some modifications. Up to conv5-2, the network is the same as VGG-16, pre-trained on ImageNet. The output of this layer is connected to two sub-models. The first one extracts context-aware features by providing a global image representation. The second one relies on a localization-type network to extract action-aware features.}
\label{FIG:LOC_TRAIN}
\end{figure}

\section{Our Method}
\label{SEC:METHODOLOGY}
Our goal is to leverage both context-aware and action-aware features for action recognition and anticipation. To this end, we introduce a multi-stage recurrent architecture based on LSTMs, depicted by Fig.~\ref{FIG:LSTM}. In this section, we first discuss our approach to extracting both feature types, and then present our multi-stage recurrent network.


\subsection{Feature Extraction}
To extract context-aware and action-aware features, we introduce the two-stream architecture shown in Fig.~\ref{FIG:LOC_TRAIN}. The first part of this network is shared by both streams and, up to conv5-2, corresponds to the VGG-16 network~\cite{VGG}, pre-trained on ImageNet for object recognition. The output of this layer is connected to two sub-models: One for context-aware features and the other for action-aware ones. We then train these two sub-models for the same task of action recognition from a single image, using a cross-entropy loss function defined on the output of each stream. In practice, we found that training the entire model in an end-to-end manner did not yield a significant improvement over training only the two sub-models. In our experiments, we therefore opted for this latter strategy, which is less expensive computationally and memory-wise. Below, we first discuss the context-aware sub-network and then turn to the action-aware one.



\subsubsection{Context-Aware Feature Extraction}
The context-aware sub-model is similar to VGG-16 from conv5-3 up to the last fully connected layer, with the number of units in the last fully-connected layer changed from 1000 (the original 1000-way ImageNet classification problem) to the number of activities $N$. 

In essence, this sub-model focuses on extracting a deep representation of the whole scene for each activity and thus incorporates context. We then take the output of its fc7 layer as our context-aware features.

\subsubsection{Action-Aware Feature Extraction}
As mentioned before, the context of an action does not always correlate with the action itself. Our second sub-model therefore aims at extracting features that focus on the action itself. To this end, we draw inspiration from the object classification work of~\cite{zhou2015learning}. At the core of this work lies the idea of Class Activation Maps (CAMs). In our context, a CAM indicates the regions in the input image that contribute most to predicting each class label. In other words, it provides information about the location of an action. Importantly, this is achieved without requiring any additional annotations.

More specifically, CAMs are extracted from the activations in the last convolutional layer in the following manner. Let \(f_l(x,y)\) represent the activation of unit \(l\) in the last convolutional layer at spatial location \((x,y)\). A score $S_k$ for each class $k$ can be obtained by performing global average pooling~\cite{NetinNetGAP} to obtain, for each unit $l$, a feature \(F^l = \sum_{x,y}{f_l(x,y)}\), followed by a linear layer with weights $\{w_l^k\}$. That is, \(S_k = \sum_k{w_l^k F_l}\). A CAM for class $k$ at location \((x,y)\) can then be computed as
\begin{equation}
M_k(x,y) = \sum_l{w_l^k f_l(x,y)}\;,
\end{equation}
which indicates the importance of the activations at location \((x,y)\) in the final score for class $k$.

\begin{figure}
\centering
\includegraphics[width=0.45\textwidth]{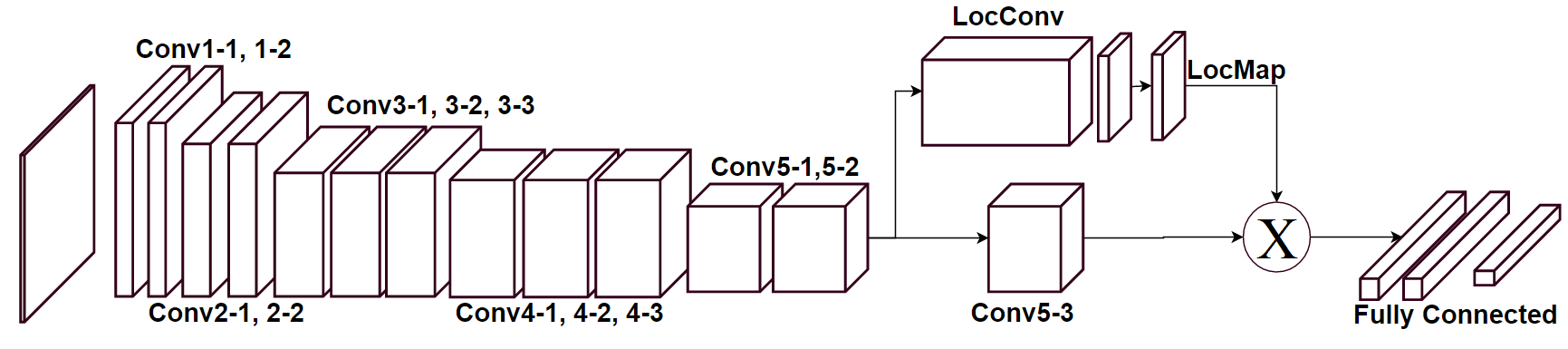}
\caption{{\bf Action-aware feature extraction.} Given the fine-tuned feature extraction network, we introduce a new layer that alters the output of conv5-3. This lets us filter out the conv5-3 features that are irrelevant, to focus on the action itself. Our action-aware features are then taken as the output of the last fully-connected layer shown here.}
\label{FIG:LOC_TEST}
\end{figure}

Here, we propose to make use of the CAMs to extract action-aware features. To this end, we use the CAMs in conjunction with the output of the conv5-3 layer of the model. The intuition behind this is that conv5-3 extracts high-level features that provide a very rich representation of the image~\cite{UnderstandingCNN} and typically correspond to the most discriminative parts of the object~\cite{DeepEdge,BuiltinFGBG}, or, in our case, the action. Therefore, we incorporate a new layer to our sub-model, whose output can be expressed as
\begin{equation}
A(x,y) = {\rm conv_{5-3}}(x,y) \times {\rm ReLU}(M_k(x,y))\;,
\end{equation}
where $\rm{ReLU}(M_c(x,y)) = max(0, M_c(x,y))$. As illustrated in Fig.~\ref{FIG:LOC_TEST}, this new layer is then followed by fully-connected ones, and we take our action-aware features as the output of the corresponding fc7 layer.

\subsection{Sequence Learning for Action Recognition}
\label{sec:seq_learn}
To effectively combine the information contained in the context-aware and action-aware features described above, we design the novel multi-stage LSTM model depicted by Fig.~\ref{FIG:LSTM}. This model first focuses on the context-aware features, which encode global information about the entire image. It then combines the output of this first stage with our action-aware features to provide a refined class prediction.

To learn this model, we introduce a novel loss function motivated by the intuition that, while we would like the model to predict the correct class as early as possible in the sequence, some actions, such as running and high jump, are highly ambiguous after seeing only the first few frames.
Ultimately, our network models long-range temporal information, and yields increasingly accurate predictions as it processes more frames. This therefore also provides us with an effective mechanism to forecast an action type given only limited input observations. Below, we discuss the two stages of our model.

%

\subsubsection{Learning Context}
The first stage of our model takes as input our context-aware features, and passes them through a layer of LSTM cells followed by a fully-connected layer that, via a softmax operation, outputs a probability for each action class. Let $\hat{y}^t_{c}(k)$ be the probability of class $k$ at time $t$ predicted by the first stage. We then define the loss for a single training sample as 
\begin{align}
\mathcal{L}_{c}(y, \hat{y}_{c}) = -\frac{1}{N}\sum^N_{k=1}\sum^T_{t=1}\Bigg[y^t(k) \log(\hat{y}^t_{c}(k)) + \nonumber \\ \frac{t(1-y^t(k))}{T}\log(1-\hat{y}^t_{c}(k))\Bigg]\;,
\end{align}
where $y^t(k)$ encodes the true activity label at time $t$, i.e., $y^t(k) = 1$ if the sample belongs to class $k$ and 0 otherwise.

This loss function consists of two terms. The first one is standard and aims at penalizing false negative with the same strength at any point in time. By contrast, the second term focuses on false positives, and its strength increases linearly over time, to reach the same weight as that on false negatives. The motivation behind this loss can be explained as follows. Early in the sequence, there can easily be ambiguities between several actions, such as running and high jump. Therefore, false positives are bound to happen, and should not be penalized too strongly. As we see more frames, however, these false positives should be encouraged to disappear. By contrast, we would like to have a high score for the correct class as early as possible. This is taken care of by the first term, which penalizes false negatives, and whose relative weight over the second term is larger at the beginning of the sequence.

\subsubsection{Learning Context and Action}
The second stage of our model aims at combining context-aware and action-aware information. Its structure is the same as that of the first stage, i.e., a layer of LSTM cells followed by a fully-connected layer to output class probabilities via a softmax operation. However, its input merges the output of the first stage with our action-aware features. This is achieved by concatenating the hidden activations of the LSTM layer with our action-aware features. We then make use of the same loss function as before, but defined on the final prediction. This can be expressed as
\begin{align}
\mathcal{L}_{a}(y, \hat{y}_{{a}}) = -\frac{1}{N}\sum^N_{k=1}\sum^T_{t=1}\Bigg[y^t(k) \log(\hat{y}^t_{a}(k)) + \nonumber \\ \frac{t(1-y^t(k))}{T}\log(1-\hat{y}^t_{a}(k))\Bigg]\;,
\end{align}
where $\hat{y}^t_{a}(k)$ is the probability for class $k$ predicted by the second stage.

The overall loss of our model can then be written as
\begin{equation}
\mathcal{L} = \mathcal{L}_{c} + \mathcal{L}_{a}\;.
\label{eq.lstm.overall}
\end{equation}
To learn the parameters of our model, we then average this loss over the training samples in a mini-batch. 

At inference time, the input RGB frames are forward-propagated though this model. We therefore obtain a probability vector for each class at each frame. While one could simply take the probabilities in the last frame to obtain the class label, via an $argmax$ operation, we propose to increase robustness by leveraging the predictions of all the frames. To this end, we make use of an average pooling of these predictions over time.

\section{Experiments}
\label{SEC:EXPERIMENTS}
In this section, we first compare our method with state-of-the-art techniques on the task of action recognition, and then analyze various aspects of our model. For our experiments, we make use of the standard UCF-101~\cite{UCF101} and JHMDB-21~\cite{JHMDB} benchmarks. The UCF-101 dataset consists of 13,320 videos of 101 action classes including a broad set of activities such as sports, musical instruments, and human-object interaction, with an average length of 7.2 seconds. UCF-101 gives a large diversity in terms of actions and with the presence of large variations in camera motion, cluttered background, illumination conditions, etc, is one of the most challenging data sets.
The JHMDB-21 dataset is another challenging dataset of realistic videos from various sources, such as movies and web videos, containing 928 videos and 21 action classes.

\paragraph{Implementation details.}
To fine-tune the network on these datasets, we used a number of data augmentation techniques, so as to reduce the effect of over-fitting. The input images were randomly flipped horizontally and rotated by a random amount in the range -8 to 8 degrees. We then extracted crops according to the following procedure:
\begin{enumerate}
\item Compute the maximum cropping rectangle with given aspect ratio ($320/240$) that can fit within the input image.
\item Scale the width and height of the cropping rectangle by a factor randomly selected in the range $0.8$-$1$.
\item Select a random location for the cropping rectangle within the orignal input image and extract that subimage.
\item Scale the subimage to $224 \times 224$.
\end{enumerate}




After these geometric transformations, we further applied RGB channel shifting~\cite{wu2015deep}, followed by randomly adjusting image brightness, contrast and saturation.

The parameters of the CNN were found by using stochastic gradient descent with a fixed learning rate of 0.001, a momentum of 0.9, a weight decay of 0.0005, and mini-batches of size 32. To train our LSTMs, we similarly used stochastic gradient descent, but with a fixed learning rate of 0.01 and a momentum of 0.9 with mini-batch size of 32. To implement our method, we use Python and Keras~\cite{keras}.

\subsection{Comparison with the State-of-the-Art}
In Tables~\ref{tab:ucf} and~\ref{tab:jhmdb}, we compare the results of our approach to state-of-the-art methods on UCF-101 and JHMDB-21, respectively by reporting the average accuracy over the given three training and testing partitions. For this comparison to be fair, we only report the results of the baselines that do not use any other information than the RGB image and the activity label. In other words, while it has been shown that additional, hand-crafted features, such as dense trajectories and optical flow, can help improve accuracy~\cite{IDT,TrajectoryPooled,VLAD3,TwoStreamNIPS,DynamicNetwork}, our goal here is to truly evaluate the benefits of our method, not of these features. Note, however, that, as discussed in Section~\ref{sec:of}, our approach can still benefits from such features. As can be seen from the tables, our approach outperforms all these baselines on both datasets.



\begin{table}[t]
\renewcommand{\arraystretch}{1.2}
\small
\centering
\caption{Comparison with state-of-the-art methods on UCF-101, three splits averaged. To provide a fair comparison, we focus on the baselines that, as us, only use the RGB frames as input (without any other information and/or hand-crafted features). }
\label{tab:ucf}
\begin{tabular}{l c }
\hline
Method & Accuracy\\
\hline
Dynamic Image Network~\cite{DynamicNetwork} & 70.0\% \\
Dynamic Image Network + Static RGB~\cite{DynamicNetwork} & 76.9\% \\
Rank Pooling~\cite{DiscriminativeRankPooling} & 72.2\% \\
Discriminative Hierarchical Ranking~\cite{DiscriminativeRankPooling} & 78.8\% \\
Realtime Action Recognition~\cite{RealTimeAction} & 74.4\% \\
LSTM~\cite{LSTMAction} & 74.5\% \\
LRCN~\cite{LRCN} & 68.8\% \\
C3D~\cite{3DCNN} & 82.3\% \\
Spatial Stream Network~\cite{TwoStreamNIPS} & 73.0\% \\
Deep Network~\cite{LargeScaleCNN} & 65.4\% \\
ConvPool (Single frame)~\cite{BeyondAction} & 73.3\% \\
ConvPool (30 frames)~\cite{BeyondAction} & 80.8\% \\
ConvPool (120 frames)~\cite{BeyondAction} & 82.6\% \\
\hline
Ours (pLGL, AvgPool, 2048 units) & {\bf 83.3\%}  \\
\hline
Comparison to State-of-the-Art & +0.7\%   \\
\hline
\end{tabular}
\end{table}

\begin{table} [t]
\renewcommand{\arraystretch}{1.2}
\small
\centering
\caption{Comparison with state-of-the-art methods on JHMDB-21, three splits averaged. Note that while we only use RGB frames as input, both baselines use motion/optical flow information.}
\label{tab:jhmdb}
\begin{tabular}{l c}
\hline
Method & Accuracy\\
\hline
Spatial-CNN~\cite{FindingActionTubes} & 37.9\% \\
Motion-CNN~\cite{FindingActionTubes} & 45.7\% \\
Full Method~\cite{FindingActionTubes} & 53.3\% \\
\\
Actionness-Spatial~\cite{actionness} & 42.6\% \\
Actionness-Temporal~\cite{actionness} & 54.8\% \\
Actionness-Full Method~\cite{actionness} & 56.4\% \\
\hline
 Ours (pLGL, AvgPool, 2048 units) & \bf{58.3\%} \\
\hline
Comparison to State-of-the-Art & +1.9\% \\
\hline

\end{tabular}
\end{table}

\subsection{Analysis}
In this section, we analyze several aspects of our method in more detail, such as the importance of each feature type, the influence of the LSTM architecture and of the number and order of the input frames. Finally, we show the effectiveness of our approach at tackling the task of action anticipation, and study how optical flow can be employed to further improve our results. All the analytical experiments were conducted on the first split of UCF-101 dataset.  

In the following analysis, we also evaluate the effectiveness of different losses.  In particular, we make use of the standard cross-entropy ({\bf CE}) loss, which only accounts for one activity label for each sequence (the activity label at time $T$). This loss can be expressed as
\begin{align}
\mathcal{L}_{CE} = \sum^{N}_{k=1}[y^T(k)\log(\hat{y}^{T}(k))\ \nonumber \\+ (1-y^T(k))\log(1-\hat{y}^{T}(k))]\;.
\end{align}

In~\cite{brain4Cars}, an exponentially growing loss ({\bf EGL}) was proposed to penalize errors more strongly as more frames of the sequence are observed. This loss can be written as
\begin{align}
\mathcal{L}_{EGL} = \sum^T_{t=1}-e^{-(T-t)}\sum^{N}_{k=1}[y^t(k)\log(\hat{y}^{t}(k)) \nonumber \\+ (1-y^t(k))\log(1-\hat{y}^{t}(k))]\;.
\end{align}

The main drawback of this loss comes from the fact that it does not strongly encourage the model to make correct predictions as early as possible. To address this issue, we introduce a linearly growing loss  ({\bf LGL}). This loss is defined as
\begin{align}
\mathcal{L}_{LGL} = \sum^T_{t=1}{-\frac{t}{T}}\sum^{N}_{k=1}[y^t(k)\log(y^{t}(k)) \nonumber \\ + (1-y^t(k))\log(1-\hat{y}^{t}(k))].
\end{align}

As shown in the experimental analysis, the linearity of this loss makes it more effective than the EGL. Our new loss, discussed in Section~\ref{sec:seq_learn} and denoted by {\bf pLGL} below, also makes use of a linearly-increasing term. This term, however, corresponds to the false positives, as opposed to the false negatives in the LGL. Since some actions are ambiguous in the first few frames, we find more intuitive not to penalize false positives too strongly at the beginning of the sequence. Our results below seem to support this, since, for a given model, our loss typically yields higher accuracies than the LGL. 

\begin{table}
\renewcommand{\arraystretch}{1.2}
\centering
\scriptsize
\caption{Importance of the different feature types using different losses. Note that combining both types of features consistently outperforms using a single one. Note also that, for a given model, our new pLGL loss yields higher accuracies the other ones.}
\label{tab:features}
\begin{tabular}{l l c c}
\hline
Feature & Sequence  & &\\
Extraction & Learning & UCF-101 & JHMDB-21 \\
\hline
Context-Aware	& LSTM (CE) 	 	& 72.38\% 	& 43.65\% \\
Action-Aware 	& LSTM (CE) 	 	& 44.24\% 	& 50.06\% \\
Context+Action & MS-LSTM (CE) 		& 78.93\%	& 54.30\% \\
\hline
Context-Aware	& LSTM (EGL) 	 	& 72.41\% 	&  44.05\% \\
Action-Aware 	& LSTM (EGL)  	 	& 77.20\% 	&  50.18\% \\
Context+Action & MS-LSTM (EGL) 		& 80.38\%	& 57.05\\
\hline
Context-Aware	& LSTM (LGL)  	& 72.58\% 	&  44.72\% \\
Action-Aware 	& LSTM (LGL)  	& 77.63\% 	&  50.34\% \\
Context+Action & MS-LSTM (LGL) 	& 81.27\%	&  57.70\% \\
\hline
Context-Aware	& LSTM (pLGL)  	& 72.71\% 	&  44.93\% \\
Action-Aware 	& LSTM (pLGL)  	& 77.86\% 	&  51.00\% \\
Context+Action & MS-LSTM (pLGL) & 83.37\%	&  58.41\% \\
\hline
\end{tabular}
\end{table}

\subsubsection{Importance of the Feature Types}
We first evaluate the importance of the different feature types, context-aware and action-aware, on recognition accuracy. To this end, we compare models trained using each feature type individually with our model that uses them jointly. For all models, we made use of LSTMs with 1024 units. Recall that our approach relies on a multi-stage LSTM, which we denote by \emph{MS-LSTM}. The results of this experiment for different losses are reported in Table~\ref{tab:features}. These results clearly evidence the importance of using both feature types, which consistently outperforms using individual ones in all settings.

\subsubsection{Influence of the LSTM Architecture}
Our second set of experiments studies the importance of using a multi-stage LSTM architecture and the influence of the number of units in our MS-LSTM. For the first one of these analyses, we compare our MS-LSTM with a single-stage LSTM that takes as input the concatenation of our context-aware and action-aware features. Furthermore, to study the importance of the feature order, we also compare our approach with a model that first processes the action-aware features and, in a second stage, combines them with the context-aware ones. Finally, we also evaluate the use of two parallel LSTMs whose outputs are merged by concatenation and then fed to a dense layer distributed over time. The results of this comparison are provided in Table~\ref{tab:LSTMArch}. Note that both multi-stage LSTMs outperform the single-stage one and the two parallel LSTMs, thus indicating the importance of treating the two types of features sequentially. Interestingly, processing context-aware features first, as we propose, yields higher accuracy than considering the action-aware ones at the beginning. This matches our intuition that context-aware features carry global information about the image and will thus yield noisy results, which can then be refined by exploiting the action-aware features.

\begin{table}
\renewcommand{\arraystretch}{1.2}
\centering
\small
\caption{Comparison of our multi-stage LSTM model with diverse fusion strategies. We report the results of simple concatenation of the context-aware and action-aware features, their use in two parallel LSTMs with late fusion, and swapping their order in our multi-stage LSTM, i.e., action-aware first, followed by context-aware. Note that multi-stage architectures yield better results, with the best ones achieved by using context first, followed by action, as proposed in this paper.}
\label{tab:LSTMArch}
\begin{tabular}{l l c}
\hline
Feature & Sequence & \\
Order & Learning & Accuracy \\
\hline
Concatenation 	& LSTM (pLGL) 				& 77.16\% \\
Swapped 		& LSTM (pLGL) 				& 78.80\% \\
Parallel 		& 2 Parallel LSTMs (pLGL) 	& 78.63\% \\
Ours 			& MS-LSTM (pLGL) 			& 83.37\% \\
\hline
\end{tabular}
\end{table}

Based on our experiments, we found that for large datasets such as UCF-101, the 512 hidden units that some baselines use (e.g.~\cite{LRCN,LSTMAction}) do not suffice to capture the complexity of the data. Therefore, to study the influence of the number of units in the LSTM, we evaluated different versions of our model with 512, 1024, 2048, and 4096 hidden units and trained the model with 80\% training data and validated on the remaining 20\%. For a single LSTM, we found that using 1024 hidden units performs best. For our multi-stage LSTM, using 2048 hidden units yields the best results. We also evaluated the importance of relying on average pooling in the LSTM. The results of these different versions of our MS-LSTM framework are provided in Table~\ref{tab:AvgPool}. This shows that, typically, more hidden units and average pooling can improve accuracy slightly.

\begin{table}
\renewcommand{\arraystretch}{1.2}
\centering
\scriptsize
\caption{Evaluating the effectiveness of applying average pooling on the softmax probabilities of all time-steps for each sample and the number of hidden units in different losses of our multi-stage LSTM model. Experiments are conducted on the first split of UCF-101 and JHMDB-21.}
\label{tab:AvgPool}
\begin{tabular}{l  c c c c}
\hline
  & Average  & Hidden & & \\
Setup  & Pooling & Units & UCF-101 & JHMDB-21\\
\hline
Ours (CE)& wo/ & 1024 & 77.26\%	& 52.80\% \\
Ours (CE)& wo/ & 2048 & 78.09\%	& 53.43\% \\
Ours (CE)& w/ & 2048 &	78.93\% & 54.30\%\\
\\
Ours (EGL)& wo/ & 1024 & 79.10\%	& 55.33\% \\
Ours (EGL)& wo/ & 2048 & 79.41\%	& 56.12\% \\
Ours (EGL)& w/ & 2048 & 80.38\%	& 57.05\%\\
\\
Ours (LGL)& wo/ & 1024 & 79.76\%	& 55.70\% \\
Ours (LGL)& wo/ & 2048 & 80.10\%	& 56.83\% \\
Ours (LGL)& w/ & 2048 & 81.27\%	& 57.70\%\\
\\
Ours (pLGL)& wo/ & 1024 & 81.94\%	& 56.24\% \\
Ours (pLGL)& wo/ & 2048 & 82.16\%	& 57.92\%\\
Ours (pLGL)& w/ & 2048 & 83.37\%	& 58.41\%\\
\hline
\end{tabular}
\end{table}

\subsubsection{Influence of the Input Frames}
As mentioned before, our model makes use of $K$ frames as input. Here, we therefore study the influence of the value $K$ and of the choice of frames on our results. To this end, we varied $K$ in the range $K=\{10,...,50\}$. For each value, we then took either the first $K$ frames of the sequence, or $K$ randomly sampled ones to our model at both training and inference time. The results for the different values of $K$ and frame selection strategies are shown in Table~\ref{tab:InputFrame}. In essence, after $K=30$, the gap between the different results becomes very small. These results also show that, while slightly better accuracies can be obtained by random sampling, using the first $K$ frames remains reliable, again, particularly for $K \geq 30$. This will be particularly important for action anticipation, where one only has access to the first frames.

\begin{table}
\centering

\caption{Influence of the Input Frames on activity recognition accuracy on UCF-101 dataset split 1.}
\label{tab:InputFrame}
\begin{tabular}{l  c c c}
\hline
Setup  & K	& First K & Sampling K \\
\hline
Ours 	& 10 & 77.4\% & 80.0\%\\
Ours	& 20 & 78.5\% & 81.2\%\\
Ours	& 30 & 81.9\% & 82.0\%\\
Ours	& 40 & 82.2\% & 82.3\%\\
\hline
Ours	& 50 & 83.4\% & 83.2\%\\
\hline
\end{tabular}
\end{table}

\subsubsection{Action Anticipation}
\begin{figure}
\centering
\includegraphics[width=0.45\textwidth]{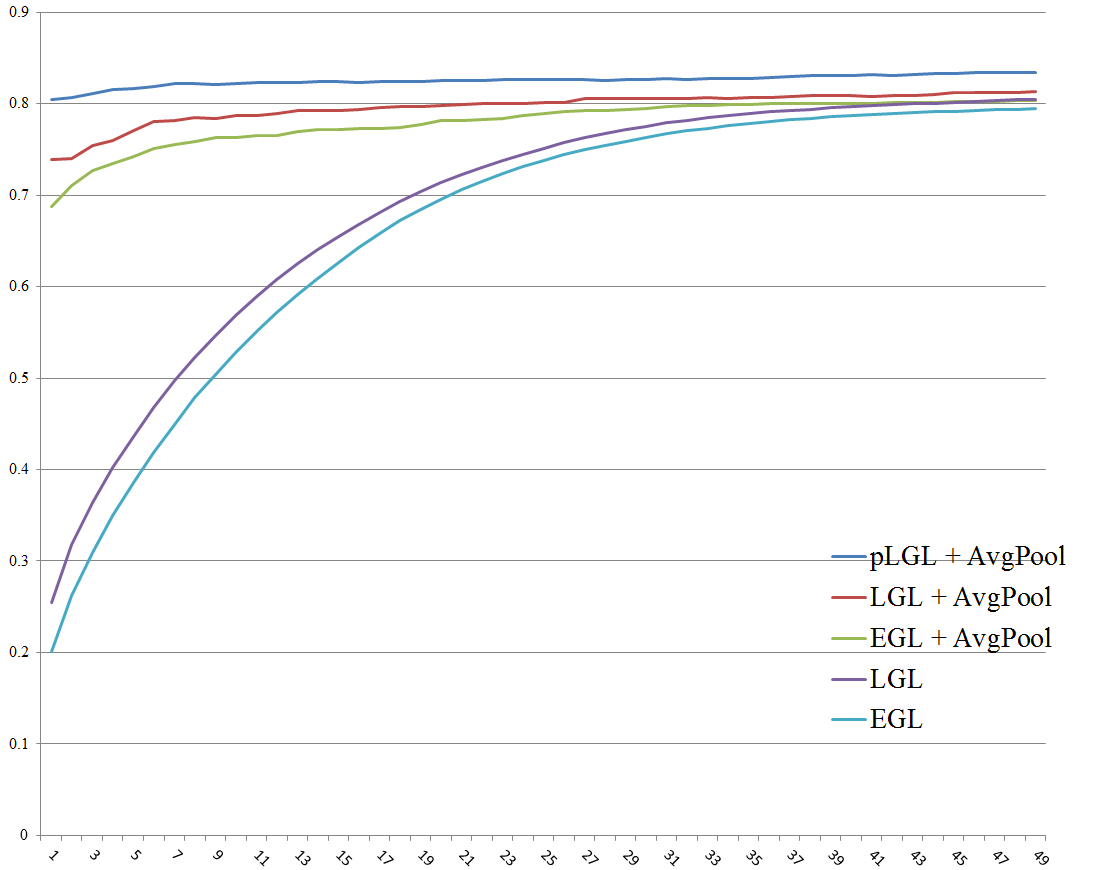}
\caption{{\bf Action Anticipation.} Evaluating the performance of different losses on anticipating the activity given partial sequential information ${x_1, ..., x_t}, t<K$ at time-step $t$. }
\label{fig:anticipation}
\end{figure}

Unlike action recognition, where the full sequence can be employed to predict the activity type, action anticipation aims at providing a class prediction as early as possible. In Fig.~\ref{fig:anticipation}, we plot the accuracy of our model as a function of the number of observed frames for different losses and with/without average pooling over time of the softmax probabilities. In this experiment, all the models were trained from sequences of $K=50$ frames. These plots clearly show the importance of using average pooling, which provides more robustness to the prediction. More importantly, they also evidence the benefits of our novel loss, which was designed to encourage correct prediction as early as possible, over other losses for action anticipation. 

\subsubsection{Exploiting Optical Flow}
\label{sec:of}
In the past, several methods have proposed to rely on optical flow to encode temporal information~\cite{TSN,TwoStreamNIPS,CNN2Stream,BeyondAction,LSTMAction}. Here, we show that our approach can also benefit from this additional source of information. To extract optical flow features, we made use of the pre-trained temporal network of~\cite{TwoStreamNIPS}. We then computed the CNN features from a stack of 20 optical flow frames (10 frames in the X-direction and 10 frames in the Y-direction), from $t-10$ to $t$ at each time $t$. As these features are potentially loosely related to action (by focusing on motion), we merge them with the input to the second stage of our multi-stage LSTM. In Table~\ref{tab:opticalFlow}, we compare the results of our modified approach with state-of-the-art methods also exploiting optical flow. Note that we consistently outperform these baselines, with the exception of those incorporating optical flow as input to two-stream network, thus learning how to exploit it and how to fuse it with RGB information. Designing an architecture that jointly leverages these motion-aware features with our context- and action-aware ones will be the topic of our future research.

\begin{table}
\renewcommand{\arraystretch}{1.2}
\centering
\small
\caption{Comparison with the state-of-the-art approaches that use optical flow features. To provide a fair comparison, we focus on the baselines that, as us, only use the RGB frames and optical flow as input (without any other hand-crafted features).}
\label{tab:opticalFlow}
\begin{tabular}{l c }
\hline
Method & Accuracy \\
\hline
Spatio-temporal ConvNet~\cite{LargeScaleCNN}			& 65.4\% \\
LRCN~\cite{LRCN} 										& 82.9\% \\
Two-Stream ConvNet~\cite{TwoStreamNIPS} 					& 88.0\% \\
VLAD3 + Optical Flow~\cite{VLAD3} 						& 84.1\% \\
Two-Stream Conv.Pooling~\cite{BeyondAction}				& 88.2\% \\
LSTM~\cite{LSTMAction} 									& 84.3\% \\
CNN features + Optical Flow~\cite{TwoStreamNIPS} 		& 73.9\% \\
ConvPool (30 frames) + OpticalFlow~\cite{BeyondAction} 	& 87.6\%\\
ConvPool (120 frames) + OpticalFlow~\cite{BeyondAction} & 88.2\%\\
\hline
Two-Stream Net Fusion~\cite{CNN2Stream}					& 92.5\% \\
TSN~\cite{TSN} 											& 93.5\% \\
\hline
Ours (pLGL, AvgPool, 2048 units) + Optical Flow 	& 88.7\% \\
\hline
\end{tabular}
\end{table}

\section{Conclusion}
\label{SEC:CONCLUSION}
In this paper, we have proposed to leverage both context- and action-aware features for action recognition and anticipation. The first type of features provides a global representation of the scene, but may suffer from the fact that some actions can occur in diverse contexts. By contrast, the second feature type focuses on the action itself, and can thus not leverage the information provided by the context. We have therefore introduced a multi-stage LSTM architecture that effectively combines these two sources of information. Furthermore, we have designed a novel loss function that encourages our model to make correct prediction as early as possible in the input sequence, thus making it particularly well-suited to action anticipation. Our experiments have evidenced the importance of our feature combination scheme and of our loss function using two standard benchmarks. Among the methods that only rely on RGB as input, our approach yields state-of-the-art action recognition accuracy. In the future, we intend to study new ways to incorporate additional sources of information, such as dense trajectories and human skeletons in our framework.


{\small
\bibliographystyle{ieee}
\bibliography{egbib}
}

\end{document}